\documentclass[lettersize,journal]{IEEEtran}
\usepackage{amsmath,amsfonts}
\usepackage{algorithmic}
\usepackage{algorithm}
\usepackage{array}
\usepackage[caption=false,font=normalsize,labelfont=sf,textfont=sf]{subfig}
\usepackage{textcomp}
\usepackage{stfloats}
\usepackage{url}
\usepackage{verbatim}
\usepackage{graphicx}
\usepackage{cite}

\usepackage{multirow}
\usepackage{amssymb}
\usepackage{booktabs}

\usepackage{times}
\usepackage{epsfig}
\usepackage{graphicx}
\usepackage{amsmath}
\usepackage{amssymb}

\usepackage{pifont}
\usepackage{booktabs}
\usepackage{multirow}
\usepackage{capt-of}
\usepackage{makecell}
\usepackage{colortbl}
\usepackage{dsfont}
\usepackage{comment}
\usepackage[dvipsnames]{xcolor}

\newcommand{\revised}[1]{{\textcolor{black}{#1}}}

\begin{document}

\title{Sharpness-aware Dynamic Anchor Selection for Generalized Category Discovery }

\author{Zhimao~Peng,~Enguang~Wang, Fei Yang, Xialei~Liu,~\IEEEmembership{Member,~IEEE}, \\ and~Ming-Ming~Cheng,~\IEEEmembership{Senior Member,~IEEE}

}

\markboth{Journal of \LaTeX\ Class Files}%
{Shell \MakeLowercase{\textit{et al.}}: A Sample Article Using IEEEtran.cls for IEEE Journals}


\maketitle

\begin{abstract}

Generalized category discovery (GCD) is an important and challenging task in open-world learning. Specifically, given some labeled data of known classes, GCD aims to cluster unlabeled data that contain both known and unknown classes. Current GCD methods based on parametric classification adopt the DINO-like pseudo-labeling strategy, where the sharpened probability output of one view is used as supervision information for the other view. However, large pre-trained models have a preference for some specific visual patterns, resulting in encoding spurious correlation for unlabeled data and generating noisy pseudo-labels. To address this issue, we propose a novel method, which contains two modules: Loss Sharpness Penalty (LSP) and Dynamic Anchor Selection (DAS). LSP enhances the robustness of model parameters to small perturbations by minimizing the worst-case loss sharpness of the model, which suppressing the encoding of trivial features, thereby reducing overfitting of noise samples and improving the quality of pseudo-labels. Meanwhile, DAS selects representative samples for the unknown classes based on KNN density and class probability during the model training and assigns hard pseudo-labels to them, which not only alleviates the confidence difference between known and unknown classes but also enables the model to quickly learn more accurate feature distribution for the unknown classes, thus further improving the clustering accuracy. Extensive experiments demonstrate that the proposed method can effectively mitigate the noise of pseudo-labels, and achieve state-of-the-art results on multiple GCD benchmarks. 

\end{abstract}


\section{Introduction}

Deep neural networks have achieved outstanding performance on visual recognition tasks \cite{he2016deep}, but this success relies heavily on the close-set assumption: both test and training images belong to fixed and known categories. However, in the open world, novel categories constantly emerge, which bring significant challenges for secure model deployment. Meanwhile, manually labeling these unknown classes and training new models is impractical for many real-world applications, so there is an urgent need for deep learning models that can automatically identify novel classes from unlabeled data.

\begin{figure}
    \centering
    \includegraphics[width=0.99\columnwidth]{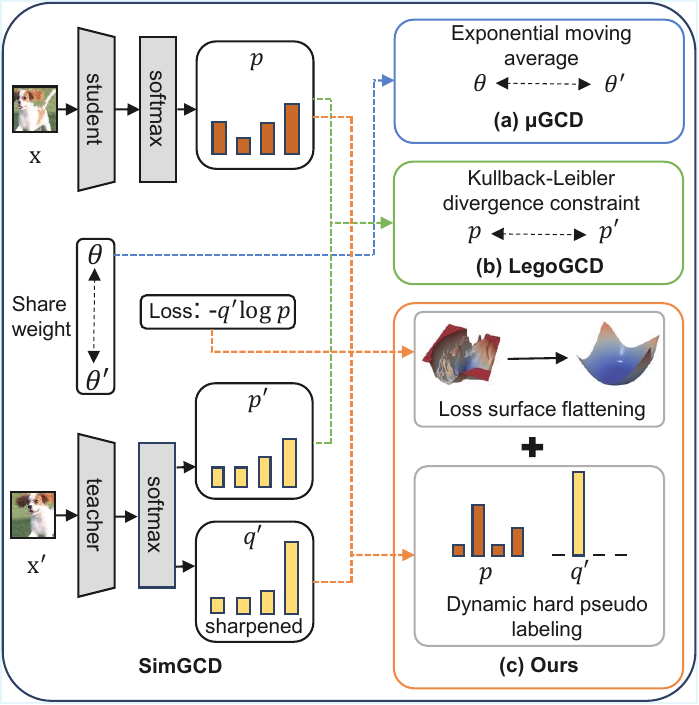}
    \caption{For parametric GCD method SimGCD~\cite{wen2023parametric} (as shown above), large pre-trained models have a preference for some visual specific patterns, resulting in encoding spurious correlation for unlabeled data and generating noisy pseudo-labels. In order to alleviate the noise of pseudo-labels, $\mu$GCD\cite{vaze2024no} introduces the mean-teacher strategy, so that the teacher is updated by the exponential moving average of the parameters of the student (as shown in (a)). LegoGCD\cite{cao2024solving} explicitly constrains the prediction probability distribution of the two views to be consistent (as shown in (b)). Unlike these ``post-hoc" smoothing methods, our method directly suppresses the generation of noisy pseudo-labels by introducing loss surface flattening and dynamic hard pseudo labeling (as shown in (c)), thus improving the cluster accuracy.   }
    \label{fig:motivation}
\end{figure}

Recently, generalized category discovery (GCD) \cite{vaze2022generalized} has been formalized to solve the above problem. Specifically, given labeled data of known classes, unlike open-set recognition\cite{Open-Set-Prompt-Tuning, Open-Set-Semi-Supervised, Open-Set-Domain-Adaptation}, which recognizes known classes while rejecting unknown classes, GCD must not only accurately classify the unlabeled data of known classes, but also cluster the unlabeled data of unknown classes according to their true semantic categories. As a practical and challenging task, GCD has attracted extensive attention, leading to numerous proposed methods \cite{vaze2022generalized,fei2022xcon,dccl,zhang2023promptcal,zhao2023learning,chiaroni2023parametric,wen2023parametric,vaze2024no,wang2024sptnet,ijcai2024p532,choi2024contrastive,cao2024solving}. In general, current GCD methods can be roughly divided into two categories: parametric and non-parametric. In non-parametric methods \cite{vaze2022generalized,fei2022xcon,dccl,zhang2023promptcal,zhao2023learning,chiaroni2023parametric,wang2024sptnet,choi2024contrastive}, deep networks only focus on learning representation that is discriminative for novel classes, while the clustering process of samples is handled by non-parametric classifiers (\textit{e.g. semi-supervised $k$-means}). In contrast, parametric methods \cite{wen2023parametric,vaze2024no,ijcai2024p532,cao2024solving} generate pseudo-labels for training data instantaneously through a parametric classification head, so that sample clustering can be completed end-to-end.

In this paper, we mainly focus on parametric methods. A pioneering parametric method, SimGCD~\cite{wen2023parametric}, as shown in Fig.\ref{fig:motivation}, employs a DINO-like self-distillation strategy for pseudo-labeling of samples, using the sharpened probability output of one view as supervision information for another \cite{caron2021emerging}. Additionally, an entropy regularization term is introduced to prevent the model from focusing excessively on known classes. Despite the encouraging performance achieved, large pre-trained models have a preference for some visual specific patterns \cite{vaze2024no}, resulting in encoding spurious correlation for unlabeled data and generating noisy pseudo-labels. To address this, 
$\mu$GCD~\cite{vaze2024no} replaces SimGCD's weight-sharing strategy with a mean-teacher approach, where the teacher model provides pseudo-labels for the student model and is updated using the moving average of the student's weights. 
This smooth parameter update enhances the robustness of the teacher model to noise, thereby improving the quality of the generated pseudo-labels. LegoGCD~\cite{cao2024solving} also mitigates the influence of noisy pseudo-labels by constraining the prediction probability distribution of two views to be consistent. However, these methods are ``post-hoc" strategies, which only smooth the noise effect, but cannot directly suppress the over-fitting to the noise, which leads to sub-optimal generalization performance.    

To address this limitation, we propose a novel method, which largely mitigates the over-fitting to noise by introducing loss sharpness penalty (LSP) and dynamic anchor selection (DAS). Specifically, LSP imposes a small perturbation on the model weight parameters to find the worst failure case \cite{foret2021sharpnessaware, wu2020adversarial}, then performing the worst case loss sharpness minimization. By penalizing loss sharpness, the model can generate a flat loss surface and become more robust to minor input variations.
For the parametric GCD framework, a flat loss surface inhibits the model from encoding trivial features, resulting in more accurate probability distributions of samples and more reliable pseudo-label generation. However, directly applying LSP is suboptimal for the GCD task, as the novel class samples lack true class labels.
This makes the training process on the novel class data contain excessive noise, leading to inaccurate gradient calculations and thus limiting the effectiveness of penalizing sharpness. In order to further improve the clustering performance, we propose a dynamic anchor selection strategy, which selects a set of category anchors for each novel class cluster based on KNN density \cite{orava2011k} and prediction confidence at the end of each training epoch. By assigning hard pseudo-labels to these anchors for the next epoch of training, not only the confidence imbalance between known and unknown classes is alleviated, but also enables the model to quickly learn more accurate feature distributions of new classes, which effectively improves the pseudo-label quality of the new classes. 

Our contribution can be summarized as follows: (1) We introduce a loss sharpness penalty (LSP) into the parametric GCD framework, which generates more robust pseudo labels for training samples by minimizing the worst-case sharpness. (2) We propose a novel dynamic anchor selection (DAS) strategy to further improve the performance by assigning hard pseudo-labels to the selected unknown classes anchors. (3) We evaluate our method on multiple generic and fine-grained GCD benchmarks. Experimental results show that our method significantly outperforms baseline methods and achieves state-of-the-art performance.

\section{Related works}
\label{sec:formatting}

\paragraph{Novel Class Discovery}
The goal of Novel Class Discovery (NCD) is to use the knowledge learned from known classes to accurately cluster unlabeled samples of unknown classes. This task was first formally defined in Han \textit{et al.}\cite{han2019learning}. Earlier works \cite{Han2020automatically,zhong2021openmix,Zhong_2021_CVPR,zhao21novel} mainly generate binary pseudo-labels based on pairwise sample similarity and then performed sample clustering by optimizing a binary cross-entropy (BCE) loss. Subsequent works \cite{fini2021unified,yang2022divide,Li_2023_CVPR,Gu_2023_ICCV} trained a unified classification head for the model by leveraging the SwAV-based online clustering. However, \revised{NCD} assumes that unlabeled data only contains new classes, which is impractical.

\paragraph{Generalized Category Discovery} \revised{Compared to NCD, Generalized Category Discovery (GCD) presents a more realistic setting: unlabeled data may come from any known or unknown classes and the model needs to cluster them correctly. The seminal work \cite{vaze2022generalized} proposed a simple non-parametric clustering strategy: first, unsupervised contrastive learning and supervised contrastive learning are performed on all the training and labeled data, respectively, and then semi-supervised $k$-means is used to cluster the unlabeled samples. Following this, many follow-up works \cite{fei2022xcon,dccl,zhang2023promptcal,zhao2023learning,RastegarECCV2024,chiaroni2023parametric,rastegar2024learn,wang2024sptnet,choi2024contrastive} have been proposed.
Among them, some works attempt to construct pseudo-relationships for samples of unknown classes to enhance representation learning: Xcon\cite{fei2022xcon} and DCCL\cite{dccl} used unsupervised clustering methods (\textit{i.e.} $k$-means and infomap) to partition the training data into multiple subsets and performed additional supervised contrastive learning between these subsets; PromptCAL\cite{zhang2023promptcal} constructed pseudo positive relationships for unlabeled samples based on the proposed affinity embedding graphs; GPC\cite{zhao2023learning} introduced a Gaussian Mixture Model to detect the compactness and separability of clusters, so as to construct dynamic prototypes and introduce an additional prototypical contrastive loss.  Overall, these pseudo-relationships construction methods are similar in spirit to our dynamic anchor selection (DAS) strategy, both attempting to introduce supervisory information for unknown classes. However, these methods can only enhance representation learning but cannot directly assist the subsequent non-parametric clustering process. In contrast, our DAS module can help parametric GCD method to generate a better feature representation space and parametric discrimination boundary simultaneously.}


\revised{Recently, a parametric classification GCD method (named SimGCD) \cite{wen2023parametric} based on self-distillation and entropy regularization has been proposed and achieved promising results. On the basis of SimGCD, $\mu$GCD \cite{vaze2024no} used mean-teacher instead of weight sharing to reduce alleviate the noise of pseudo-labels. LegoGCD\cite{cao2024solving} alleviated the catastrophic forgetting of known classes in SimGCD during training by introducing local entropy regularization and Dual-views Kullback–Leibler divergence constraint. In contrast, our work focuses on generating flat loss surface by introducing loss sharpness penalty, so as to generating more reliable pseudo-labels. In addition, the proposed dynamic anchor selection strategy is also related to \cite{ijcai2024p532} and \cite{ma2024active}, but our method does not require multi-stage training and human intervention. More importantly, The proposed dynamic anchor selection (DAS) can cooperate with the loss sharpness penalty (LSP) to further improve the model performance.}

\paragraph{Sharpness Aware Minimization (SAM)}
SAM~\cite{foret2021sharpnessaware} and its variants \cite{kwon2021asam,zheng2021regularizing,du2022efficient,liu2022towards} simultaneously optimizes the supervised loss and the sharpness of loss landscape, allowing the model to find a flat loss surface. The flatness obtained by SAM can significantly improve generalization performance, and this advantage has been demonstrated in domain generlization~\cite{wang2023sharpness}, test-time adaptation\cite{sharpness-Test-time}, prompt learning\cite{sharpness-Prompt-Learning}, long-tailed learning~\cite{zhou2023class,zhou2023imbsam}, failure prediction~\cite{zhu2022rethinking} and semi-supervised learning~\cite{huang2023flatmatch}. However, SAM is only used to optimize the training of known class data, whether or how it can be leveraged to improve learning on unknown class data has not been explored. 

\section{Our Method}
\begin{figure*}[t]
    \centering
    \includegraphics[width=0.99\textwidth]{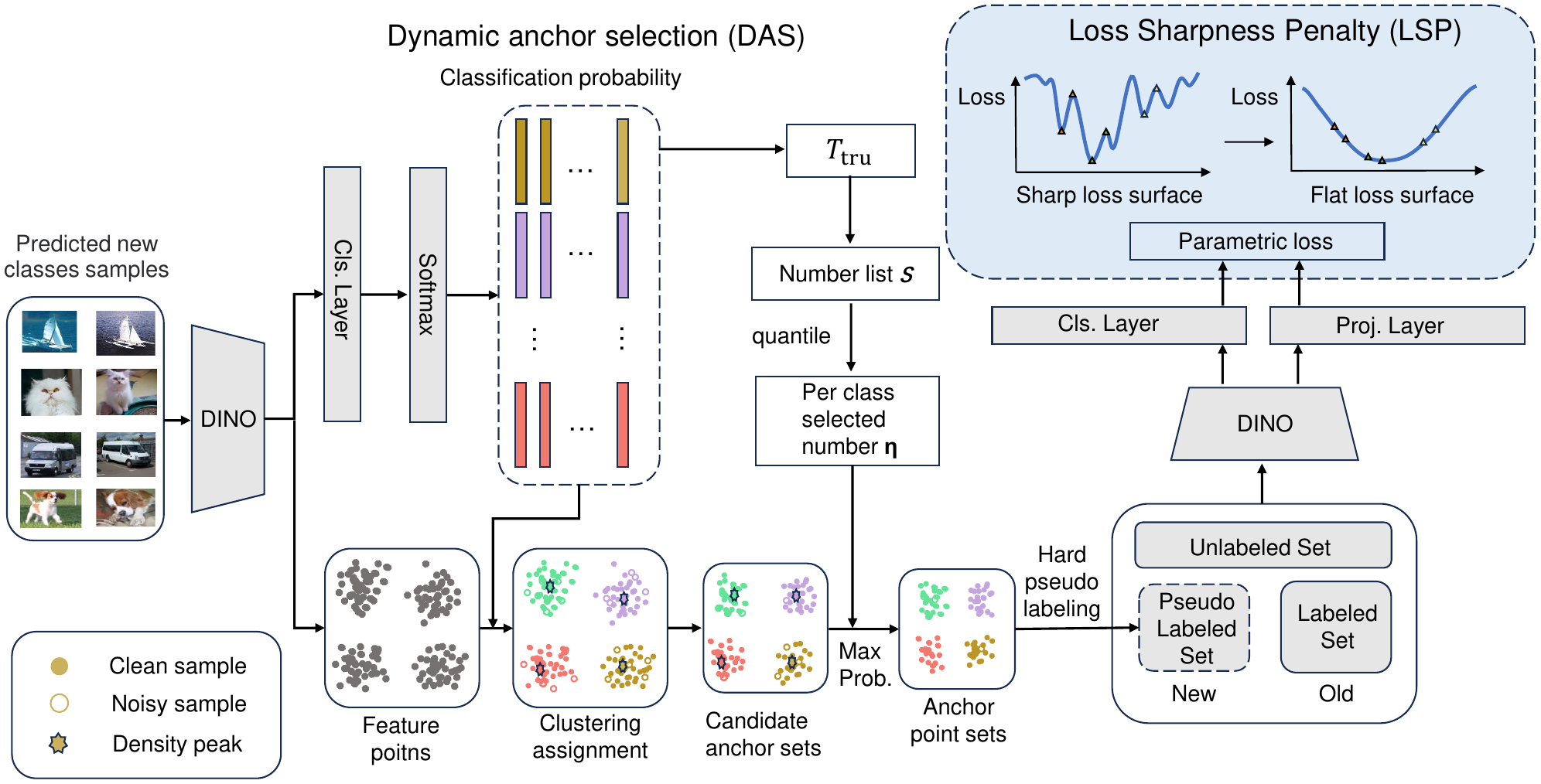}
    \caption{ The framework of the proposed method. It mainly contains two modules, LSP and DAS. LSP generates a flat loss surface by penalizing the loss sharpness, thus suppressing the overfitting of noisy samples. The DAS module selects anchor points for new classes and assigns hard pseudo-labels to them to further improve the performance. \revised{The ``clustering assignment'' of a certain sample is the index corresponding to the maximum class probability of the parametric classification head.}
    }
    \label{fig:method}
    \end{figure*}

In this section, we will introduce the proposed  method. First, we describe the problem formulation of GCD. Then we introduce the general parametric framework of GCD. Finally, we elaborate on the proposed method. 

\subsection{Problem Formulation}
Given a training dataset $\mathcal{D}$ that used for GCD task, it contains two parts: labeled set $\mathcal{D}_l$ and unlabeled set $\mathcal{D}_u$, where $\mathcal{D}_l=\left\{x_i^l, y_i^l\right\}_{i=1}^N$, with $y_i^l \in \mathcal{Y}_l$, and $\mathcal{D}_u=\left\{x_j\right\}_{j=1}^M$, with underlying category space $\mathcal{Y}_u$. For GCD task, $\mathcal{Y}_l$ and $\mathcal{Y}_u$ stand for ``Old" classes and ``All" classes respectively and $\mathcal{Y}_l \subseteq \mathcal{Y}_u$. $\mathcal{Y}_n$ stand for ``New" classes and $\mathcal{Y}_n=\mathcal{Y}_{\text {u }} \backslash \mathcal{Y}_l$. The goal of GCD is to correctly cluster samples in $\mathcal{D}_u$ according to their true semantic categories with the help of $\mathcal{D}_l$. 

\subsection{General Parametric GCD Framework}
As shown in Fig. \ref{fig:method}, following \cite{wen2023parametric,cao2024solving}, the architecture of General Parametric GCD method consists of three parts: an encoder $f(.)$ implemented by DINO pretrained ViT, a projection head $\phi$ and a classification head $h$. Given the training data $\mathcal{D}$, two randomly augmented views of an input image ($x_i, x'_i$) in a mini batch $B$ are first encoded as feature vectors ($f(x_i), f(x'_i)$) by the encoder. $z_i$ and $z'_i$ are the  $\ell_2$-normalized projection head output $\phi(f(x_i))$ and $\phi(f(x'_i))$ respectively. By feeding the $\ell_2$-normalized $f(x_i)$ and $f(x'_i)$ into the $\ell_2$-normalized $h$, the corresponding logits $l_i$ and $l'_i$ can be obtained, respectively. The probability distribution $p_i$ of $x_i$ can be obtained by feeding $l_i$ into a softmax layer $\sigma$: $\boldsymbol{p}_i=\sigma\left(\boldsymbol{l}_i / \tau_s\right)$ and the corresponding soft pseudo label of another view $x'_i$ can be defined as $\boldsymbol{q'}_i = \sigma\left(\boldsymbol{l'}_i / \tau_t\right)$,  where $\tau_s$ is a temperature and $\tau_t$ is a sharper version of $\tau_s$.

All loss functions of SimGCD \cite{wen2023parametric} can be divided into representation learning loss and parametric classification loss. For representation learning loss, it consists of self-supervised contrastive learning loss on all training data $\mathcal{D}$:

\begin{equation}
    \mathcal{L}_\phi^{u}= \frac{1}{|B|} \sum_{i \in B} - \log \frac{\exp \left(\mathbf{z}^{\top}_{i} \cdot \mathbf{z}^{\prime}_{i} / \tau^{u}\right)}{\sum_{i }^{i\neq j} \exp \left(\mathbf{z}^{\top}_{i} \cdot \mathbf{z}_{j} / \tau^{u}\right)},
\end{equation}
and supervised contrastive loss on all labelled data $\mathcal{D}_l$:

\begin{equation}
    \mathcal{L}_\phi^{s}=\frac{1}{\left|B^L\right|} \sum_{i \in B^L} \frac{1}{|\mathcal{P}(\mathbf{x})|} \sum_{\mathbf{z}^{+}_{i} \in \mathcal{P}(\mathbf{x})} -\log \frac{\exp \left(\mathbf{z}^{\top}_{i}  \mathbf{z}^{+}_{i} / \tau^{s}\right)}{\sum_{i }^{i\neq j} \exp \left(\mathbf{z}^{\top}_{i}  \mathbf{z}_{j} / \tau^{s}\right)},
\end{equation}
where $\tau^{u}$ and $\tau^{s}$ are temperature values. $B^L$ is labeled part of $B$. $\mathcal{P}(\mathbf{x})$ is the positive sample set of $z_i$. \revised{$\mathbf{z}_i^{+}$ is a positive sample of $z_i$.} The representation learning can be written as $\mathcal{L}_{rep}=(1-\lambda) \mathcal{L}_{\phi}^u+\lambda \mathcal{L}_{\phi}^s$.

For parametric classification loss, it consists of a cross-entropy loss on all labeleld data $\mathcal{D}_l$:
\begin{equation}
    \mathcal{L}_{ce}^{s}= \frac{1}{\left|B^L\right|} \sum_{i \in B^L} \ell\left(\boldsymbol{y}_i, \boldsymbol{p}_i\right),
\end{equation}
and a self-distillation loss on all training data $\mathcal{D}$:

\begin{equation}
\mathcal{L}_{\mathrm{ce}}^u=\frac{1}{|B|} \sum_{i \in B} \ell\left(\boldsymbol{q}_i^{\prime}, \boldsymbol{p}_i\right)-\varepsilon H(\boldsymbol{p}_{avg}),
\end{equation}
where $H(\boldsymbol{p}_{avg})$ is a mean-entropy maximum regularization term which used to prevent trivial solutions:

\begin{equation}
H(\boldsymbol{p}_{avg})= -\sum_c \boldsymbol{p}^{c}_{avg} \log \boldsymbol{p}^{c}_{avg},
\end{equation}
where $\boldsymbol{p}_{avg}=\frac{1}{2|B|} \sum_{i \in B}\left(\boldsymbol{p}_i+\boldsymbol{p}_i^{\prime}\right)$ is the mean probability distribution of one mini-batch, $c$ is all the class index. The parametric classification loss can be written as $\mathcal{L}_{\text {cls}}=(1-\lambda) \mathcal{L}_{\text {ce}}^u+\lambda \mathcal{L}_{\text {ce}}^s$. Therefore, the final loss function $\mathcal{L}_{ALL}$ can be represented as $\mathcal{L}_{ALL} = \mathcal{L}_{\text {rep }}+\mathcal{L}_{\mathrm{cls}}$.

In summury, the key in SimGCD's success lies in self-distillation and mean-entropy maximum regularization. For the former, by using the sharpened prediction of one view as supervision information for the other view of the same sample, model establishes similarities and dissimilarities between different views, and then clusters similar samples together. The latter encourages the output of the model to tend to diverse probability distributions, thus avoiding the probability distribution of samples concentrated in a certain class. Despite the encouraging results of SimGCD, current framework suffer from representation bias: since the feature encoder is pre-trained on the large-scale dataset, it makes the model have a preference for some specific visual patterns~\cite{vaze2024no}. This makes the model encode spurious correlation for some samples during the training process of GCD, thus generating inaccurate pseudo-labels for them. Although $\mu$GCD~\cite{vaze2024no} and LegoGCD~\cite{cao2024solving} adopt mean-teacher and dual-view alignment strategies to alleviate this problem respectively, these methods only smooth the model fitting to noisy pseudo-labels, but cannot directly suppress the generation of noisy pseudo-labels.      

\subsection{Loss Sharpness Penalty (LSP)} 
\label{subsec:LSP}
To mitigate spurious correlations caused by mismatched feature transfer \cite{vaze2024no}, it is helpful to encourage the model to perform well on a wide range of weight configurations, as this prevents the model from relying too much on specific visual patterns, thereby facilitating the more accurate feature encoding of the sample. Specifically, we use the largest loss change when a small perturbation is applied to the model as a measure of sharpness \cite{foret2021sharpnessaware, wu2020adversarial}. The intuition is that finding minima in flat regions 
 (\textit{i.e.} regions of low sharpness) of the loss surface will make the model less sensitive to small perturbations or noise in the input data. In contrast, sharp minima (\textit{i.e.} regions of high sharpness) indicate that the loss is very sensitive to small changes of the weight parameters, which can lead to over-fitting of the model to noisy samples. Therefore, we encourage the model to generate a flat loss surface during training. Specifically, we minimize the training loss while also penalizing the sharpness of the loss.  Specifically, given the overall loss function of the model $\mathcal{L}_{ALL}(\theta)$, where $\theta$ is the model weight parameters. We maximize the empirical risk by applying a small perturbation $\epsilon$ to the model parameters $\theta$. In order to find this worst failure case, during the model training, the gradient of $\mathcal{L}_{ALL}(\theta)$ is first calculated and then the parameters $\theta$ are moved to the parameter points that amplify the loss in the direction of this gradient. The perturbed weight parameters (\textit{i.e.} worst case point) $\theta_{\text {pert }}$ can be expressed as follows:

\begin{equation}
\begin{aligned}
\theta_{\text {pert }}
&=\theta+\operatorname{argmax}_{\|\epsilon\|_p \leq \rho} \mathcal{L}_{ALL}(\theta+\epsilon) \\
&\approx \theta+ \underset{\|\epsilon\|_p \leq \rho}{\arg \max } \epsilon^{\top} \nabla_\theta \mathcal{L}_{ALL}(\theta) \\
&\stackrel{p=2}{\approx} \theta+ \rho \frac{\nabla_\theta \mathcal{L}_{ALL}(\theta)}{\left\|\nabla_\theta \mathcal{L}_{ALL}(\theta)\right\|_2}, 
\end{aligned}
\end{equation}
where $\rho$ is a hyperparameter that control the amplitude of the perturbation. In order to perform the worst-case loss sharpness penalty, we can then use the perturbed parameters $\theta_{\text {pert }}$ to update the original model parameters $\theta$ by computing the gradient of the loss function $\mathcal{L}_{ALL}\left(\theta_{\text {pert }}\right)$:

\begin{equation}
\theta \leftarrow \theta-\delta \nabla_\theta \mathcal{L}_{ALL}\left(\theta_{\text {pert }}\right),
\end{equation} 
where $\delta$ is the learning rate. By performing the loss sharpness minimization described above, the model can produce a flatter loss surface, thereby suppressing the encoding of trivial features.

\subsection{Dynamic Anchor Selection}
\label{subsec:DAS}
 For parametric GCD methods, although LSP is expected to help generate more robust pseudo-labels by reducing encoding spurious correlation, directly applying it to the original loss function would lead to sub-optimal performance. This is because for the GCD task, the novel classes do not have any true class labels, which makes the pseudo-labels of the novel classes more noisy during training. When LSP finds the worst failure case, the novel classes component of the loss gradient is mainly provided by the novel classes component of the self-distillation loss gradient. Noisy novel class pseudo-labels can lead to erroneous novel classes gradient computations, thus impeding the effectiveness of penalizing loss sharpness. In order to further improve the model performance, we propose to select ``anchor points" for the new classes during training. These anchor points are defined as samples that represent the semantic category of that cluster. By assigning one-hot hard pseudo-label to these anchor points, the model can quickly learn a more accurate feature distribution for the new class, which not only alleviates the confidence imbalance between the new classes and the old classes, but also calibrates the feature encoding of the pre-trained model for the new classes samples, thereby improving the accuracy of the generated pseudo-labels. 

Inspired by \cite{ijcai2024p532,ma2024active}, we use the clustering results of SimGCD with LSP \revised{(\textit{i.e.} ``clustering assignment'' in Fig.\ref{fig:method} )} for the anchor selection in the initial stage and use them for training in the beginning $\alpha$ epochs. After that, the clustering results of the previous of our method epoch is used for anchor selection of current epoch. Specifically, given the predicted probability distribution $P=\left\{p_1, p_2, \ldots, p_m\right\}$ of all assigned new class samples $X=\left\{x_1, x_2, \ldots, x_m\right\}$ and the sample clusters of novel classes $\mathcal{C}=\left\{\mathcal{C}_j:j \in(1, \ldots, \left|\mathcal{Y}_n\right|) \right\}$, where $\mathcal{C}_j=\left\{\left(\boldsymbol{x}_i, p_i\right)\right\}_{i=1}^{|\mathcal{C}_j|}$, $|\mathcal{C}_j|$ is the sample number of $\mathcal{C}_j$, we first define a truncation threshold:

\begin{equation}
 T_{tru} = p_{avg} + (p_{max} - p_{avg})\times\omega,
\end{equation}
where $p_{avg} = \frac{1}{m} \sum_{i=1}^m \max (p_i)$ is the average of class confidence of $X$, $p_{max} = \max(P)$ is highest class confidence of $X$. \revised{$\omega$ is a is a weighting factor, which is used to control the value of $T_{t r u}$.}  We count the number of samples in each cluster whose confidence is higher than the truncation threshold, which can be expressed as $S=\left\{s_{1}, \ldots, s_{\left|\mathcal{Y}_n\right|}\right\}$, where $s_j=\sum_{i=1}^{|\mathcal{C}_j|} \mathds{1}\left[\max (p_i)>T_{tru}\right]$, $j \in\left[1, \ldots,\left|\mathcal{Y}_n\right|\right]$. We seletct a uniform number of anchors for each novel class cluster, and this number can be defined as $\eta$, whose value is determined by the $\gamma$ quantile of $S$.

Current anchor selection methods either computationally complex or require human intervention \cite{ijcai2024p532,ma2024active}, and all of them require multi-stage training. To efficiently and accurately select anchors for each cluster, we propose a novel dynamic anchor selection (DAS) strategy. Considering that the samples with high feature density values in a cluster are often representative samples, we first look for the sample with the highest feature density value in a cluster as the approximate class center of that cluster. The $K$-nearest neighbor density estimate \cite{orava2011k} can be written as follows:

\begin{equation}
d(x)=\frac{k}{n V_d\left[d_k(x)\right]^d},
\end{equation}
where $V_d$ is the volume of the unit ball in $d$-dimensional space. $d$ is the dimension of the feature. $d_k(x)$ is the Eucilidean distance from $x$ to its $k$-th nearest neighbor.

To boost the robustness to noise, following \cite{wang2022unsupervised}, we replace $d_k(x)$ with the mean of distance from $x$ to its $k$-nearest neighbors:

\begin{equation}
\hat{d}(x)=\frac{k}{n V_d\left[\, \overline{d}(x_i,k)\right]^d },
\end{equation}
where
\begin{equation}
\overline{d}(x_i,k)=\frac{1}{k} \sum_{j=1}^k d\left(x_i,x_{j(i)}\right).
\end{equation}
Since our purpose is to find the sample with the highest feature density, we only need to find the sample corresponding to the largest $1/\overline{d}(x_i,k)$. Once we find the density peak sample $x_p$, we can select $x_p$ and the other $|\mathcal{O}_j|-1$ samples closest to $x_p$ as the candidate anchor point set $\mathcal{O}_j$ for cluster $\mathcal{C}_j$, where $|\mathcal{O}_j|$ is the sample number of $\mathcal{O}_j$, $|\mathcal{C}_j|$ is the sample number of $\mathcal{C}_j$ and $|\mathcal{O}_j|=\beta|\mathcal{C}_j|$. $\beta$ is a scale factor and $\beta \in [0,1]$. Finally, we can simply use the maximum classification probability $\max(p)$ as the metric for anchor selection, that is, select $\eta$ samples from $\mathcal{O}_j$ in descending order of the value of $\max (p_i)$ as the final anchor point set $\mathcal{A}_j$ for each cluster $\mathcal{C}_j$.    

During the training time, the labeled set can be updated as $D'_l = D_l \cup \mathcal{A}$, and then $B^L$, $\mathcal{L}_\phi^s$, $\mathcal{L}_{ce}^s$ and $\mathcal{L}_{ALL}$ are updated as $B^{L'}$, $\mathcal{L}_\phi^{s'}$, $\mathcal{L}_{ce}^{s'}$ and $\mathcal{L}_{ALL}^{'}$. Therefore, the final loss sharpness penalty is performed on $\mathcal{L}_{ALL}^{'}$. 
\revised{Since LSP suppress the encoding of trivial features by generating flat loss surfaces, the parametric model can generate more accurate classification probability distributions for samples. This, in turn, promotes the effectiveness of the dynamic anchor selection based
on classification probability threshold. The corresponding experimental analysis is shown in Table \ref{tab:anchor}.}
Algorithm~\ref{alg:algorithm1} lists the pseudo-code of our method.

\begin{algorithm}[t]
    \small
	\renewcommand{\algorithmicrequire}{\textbf{Input:}}
	\renewcommand{\algorithmicensure}{\textbf{Output:}}
	\caption{Sharpness-aware Dynamic Anchor Selection for Generalized Category Discovery.}
	\label{alg:algorithm1}
	\begin{algorithmic}[1]
		\REQUIRE Labeled set $\mathcal{D}_l=\left\{x_i^l, y_i^l\right\}_{i=1}^N$, unlabeled set $\mathcal{D}_u=\left\{x_j\right\}_{j=1}^M$. 
            \REQUIRE Initial sample assignment of novel classes $\mathcal{C}$. 
            \REQUIRE Loss weight $\lambda$, scale factor $\beta$, number of anchor points per novel class $\eta$, the number of epoches trained using anchor points $\alpha$, max epochs $T_{max}$.
		\REQUIRE ViT-B-16 DINO backbone $f(.)$, projection head $\phi$, classification head $l$.
            \REQUIRE Representation learning loss $\mathcal{L}_{rep}=(1-\lambda) \mathcal{L}_{\phi}^u+\lambda \mathcal{L}_{\phi}^s$,
            parametric classification loss $\mathcal{L}_{\text {cls}}=(1-\lambda) \mathcal{L}_{\text {ce}}^u+\lambda \mathcal{L}_{\text {ce}}^s$ and final loss function $\mathcal{L}_{ALL} = \mathcal{L}_{\text {rep }}+\mathcal{L}_{\mathrm{cls}}$.
            \FOR {$t=1,2,...,T_{max}$}
		\STATE Use $\mathcal{L}_{ALL}$ with LSP to train $f$, $\phi$ and $l$ and obtain the initial cluster assignment $\mathcal{C}$.
		\ENDFOR
            \STATE Use the DAS module to obtain the final anchor set $\mathcal{A}$ from $\mathcal{C}$.
            \STATE Update the labeled set $\mathcal{D}_l$ to $D'_l = D_l \cup \mathcal{A}$, then $\mathcal{L}_\phi^s$, $\mathcal{L}_{ce}^s$ and $\mathcal{L}_{ALL}$ are updated as $\mathcal{L}_\phi^{s'}$, $\mathcal{L}_{ce}^{s'}$ and $\mathcal{L}_{ALL}^{'}$.
		\FOR {$t=1,2,...,T_{max}$}
		\IF{$t\leq \alpha$}{
			\STATE Use $\mathcal{L}_{ALL}^{'}$ with LSP to train $f$, $\phi$ and $l$.}
		\ELSE
                \STATE Use the DAS module to obtain the final anchor set $\mathcal{A}$ from $\mathcal{C}$.
                \STATE Update the labeled set $\mathcal{D}_l$ to $D'_l = D_l \cup \mathcal{A}$, then $\mathcal{L}_\phi^s$, $\mathcal{L}_{ce}^s$ and $\mathcal{L}_{ALL}$ are updated as $\mathcal{L}_\phi^{s'}$, $\mathcal{L}_{ce}^{s'}$ and $\mathcal{L}_{ALL}^{'}$.
                \STATE Use $\mathcal{L}_{ALL}^{'}$ with LSP to train $f$, $\phi$ and $l$.
		\ENDIF
		\ENDFOR
		\STATE \textbf{return} CNN backbone $f$, Classification head $l$.
	\end{algorithmic}
\end{algorithm}

\section{Experiments}

\begin{table}[h]
      \small
      \centering
      \caption{The split details of labeled set and unlabeled set in each dataset used in our experiments.}
      \vspace{-.1in}
      \resizebox{0.8\columnwidth}{!}{\begin{tabular}{lcccc}
        \toprule
        \multirow{2}[3]{*}{Dataset}  &
        \multicolumn{2}{c}{Labeled set $D_l$} & \multicolumn{2}{c}{Unlabeled set $D_u$}\\
        \cmidrule(lr){2-3} \cmidrule(lr){4-5}
         & Images & Classes & Images & Classes \\
        \midrule
        CIFAR-100 & 20K & 80 & 30K & 100 \\
        ImageNet-100 & 31.9K & 50 & 95.3K & 100 \\
        CUB & 1.5K & 100 & 4.5K & 200 \\
        Stanford Cars & 2.0K & 98 & 6.1K & 196 \\
        FGVC-Aircraft & 1.7K & 50 & 5.0K & 100 \\
        Herbarium 19 & 8.9K & 341 & 25.4K & 683 \\
        \bottomrule
      \end{tabular}}
      \label{tab:datasets}
\end{table}

\subsection{Experimental Setup}

\paragraph{Datasets}

  \begin{table*}[t]
    \centering
    \caption{Results on generic and fine-grained image recognition datasets.}
    \label{tab:all_results}
    \resizebox{\textwidth}{!}{%
    \begin{tabular}{lccc|ccc|ccc|ccc|ccc|ccc|ccc}
    \cmidrule[1pt]{1-22}
    
    \multicolumn{1}{c}{\multirow{2}{*}{Method}} & \multicolumn{3}{c}{CIFAR-100} & \multicolumn{3}{c}{ImageNet-100} & \multicolumn{3}{c}{CUB} & \multicolumn{3}{c}{Stanford Cars} & \multicolumn{3}{c}{FGVC-Aircraft} & \multicolumn{3}{c}{Herbarium 19} & \multicolumn{3}{c}{Average}  \\ 
    \cmidrule(rl){2-4} \cmidrule(rl){5-7} \cmidrule(rl){8-10} \cmidrule(rl){11-13} \cmidrule(rl){14-16} \cmidrule(rl){17-19} \cmidrule(rl){20-22}
    \multicolumn{1}{c}{}                             & All      & Old     & New     & All       & Old       & New  & All       & Old       & New & All       & Old       & New  & All       & Old       & New & All       & Old       & New & All       & Old       & New \\ \cmidrule[0.5pt]{1-22}
    k-means~\cite{arthur2007k}                          & 52.0     & 52.2    & 50.8    & 72.7      & 75.5      & 71.3  & 34.3 &38.9 &32.1 &12.8& 10.6& 13.8& 16.0& 14.4& 16.8 &  13.0 & 12.2 &  13.4     &  33.5 & 34.0 & 33.0 \\
    RS+~\cite{Han2020automatically}                          & 58.2     & 77.6    & 19.3    & 37.1      & 61.6      & 24.8  & 33.3 &51.6& 24.2& 28.3& 61.8& 12.1&
    26.9&36.4&22.2  &  27.9 & 55.8 & 12.8 & 35.3  & 57.5 & 19.2 \\
    UNO+~\cite{fini2021unified}                                & 69.5     & 80.6    & 47.2    & 70.3      & 95.0      & 57.9  &35.1& 49.0& 28.1& 35.5& 70.5 &18.6&40.3&56.4 &32.2 &  28.3 & 53.7 & 14.7   & 46.5  & 67.5 & 33.1\\
    ORCA~\cite{cao2022openworld}                                    & 69.0     &  77.4    &  52.0    & 73.5      & 92.6      & 63.9  &35.3&  45.6&  30.2&  23.5 & 50.1 & 10.7 &22.0&31.8 &17.1 &   20.9 &  30.9  &  15.5 & 40.7  & 54.7 &  31.6 \\
    GCD~\cite{vaze2022generalized}                               &	 73.0	 &    76.2     & 66.5   &   74.1&   89.8     &   66.3  & 51.3  &  56.6  &  48.7   &  39.0 & 57.6 & 29.9         &  45.0 & 41.1  & 46.9 &  35.4 &  51.0 & 27.0 & 53.0  & 62.1 & 47.6  \\
    DCCL~\cite{dccl}                               &	75.3	 &    76.8     & 70.2   &  80.5&   90.5     &   76.2  &63.5          &    60.8     &  64.9       &     43.1   & 55.7 &   36.2         &  --  &   --       &   -- & -- & -- & --&  --  & --  & --   \\
    PromptCAL~\cite{zhang2023promptcal}                               & 81.2 &  84.2 &  75.3   &  83.1 &   92.7   &  78.3 &62.9         &    64.4     &  62.1       &  50.2      & 70.1 &  40.6  &  52.2  & 52.2  & 52.3 & 37.0 & 52.0 & 28.9 & 61.1  & 69.3 & 56.3  \\
    GPC~\cite{zhao2023learning}                          & 77.9     &  85.0    &  63.0    &  76.9      & 94.3      & 71.0 &  55.4     &  58.2    & 53.1    & 42.8     &  59.2    &  32.8    &  46.3      & 42.5      & 47.9  &  -- & -- & -- & --  & -- & --
        \\ 
    $\mu$GCD~\cite{vaze2024no}                          & --     &  --    &  --    &  --      & --      & -- &  65.7     &  68.0    & 64.6    & 56.5     &  68.1   &  50.9     &  53.8       & 55.4     & 53.0  &  45.8 & \textbf{61.9}  & 37.2 & --  & --  & --
        \\ 
    InfoSieve~\cite{rastegar2024learn}                          & 78.3     &  82.2    &  70.5    &  80.5      & 93.8      & 73.8 &  \textbf{69.4}     &  \textbf{77.9}    & 65.2    & 55.7     &  74.8    &  46.4    &  56.3       & \textbf{63.7}       & 52.5  &  41.0 & 55.4 & 33.2  & 63.5  & 74.6 & 
      56.9  \\ 
    RSS~\cite{ijcai2024p532}                        & 80.7 & 81.4 & 79.1  & 85.4 & 92.8  & 81.7 &  65.1 & 66.2 & 64.5 & 55.8 & 72.2 & 47.9 &  54.7 & 58.4 & 52.8  & - & - & - & -  & -  & - 
        \\ 
    SPTNet~\cite{wang2024sptnet}                        & 81.3 & 84.3 & 75.6  & 85.4 & 93.2  & 81.4 &  65.8 & 68.8 & 65.1 & 59.0 & 79.2 & 49.3 &  \textbf{59.3} & 61.8 & \textbf{58.1}  & 43.4 & 58.7 & 35.2 & 65.7  & 74.3  & 60.8 
        \\ 
    CMS~\cite{choi2024contrastive}                         & 82.3 & \textbf{85.7} & 75.5    & 84.7 & \textbf{95.6}  & 79.2 &  68.2 & 76.5 & 64.0    & 56.9 & 76.1 & 47.6    &  56.0 & 63.4 & 52.3  & 36.4 & 54.9 & 26.4 & 64.1  & \textbf{75.4} & 57.5  
        \\ 
    LegoGCD~\cite{cao2024solving}                          & 81.8  & 81.4  & \textbf{82.5}    &  86.3 & 94.5 & 82.1  &  63.8 & 71.9 & 59.8    & 57.3  & 75.7 & 48.4    & 55.0 & 61.5 & 51.7   &  45.1  & 57.4  & 38.4 & 64.9  & 73.7  & 60.5
        \\ 
    \midrule
    SimGCD~\cite{wen2023parametric}   & 80.1     & 81.2    & 77.8    & 83.0      & 93.1      & 77.9 & 60.3 & 65.6 &
    57.7 & 53.8 & 71.9 & 45.0 & 54.2 & 59.1 & 51.8  & 44.0 & 58.0 & 36.4 & 62.6  & 71.5 & 57.8 \\
    Ours   &  \textbf{83.7}  & 84.6 & 82.0 & \textbf{87.1} & 93.9 & \textbf{83.7} & 69.0  & 69.1  & \textbf{68.9} & \textbf{65.5} & \textbf{79.7} & \textbf{58.8} & 55.6
 & 59.2& 53.8 & \textbf{49.6} 
 & \textbf{61.9} & \textbf{43.0} & \textbf{68.4}  & 74.7 & \textbf{65.0}
         \\\cmidrule[1pt]{1-22}
    \end{tabular}
    
    }
    \end{table*}

We evaluate the proposed method on six datasets, which involve two generic datasets (including CIFAR-100 and ImageNet-100), three balanced fine-grained datasets from SSB (including CUB, Stanford Cars, and FGVC-Aircraft), and a long-tailed dataset fine-grained dataset Herbarium 19. For each dataset, following previous works~\cite{vaze2022generalized,wen2023parametric}, 
we use 50\% of the images sampled from training set of known classes as the labeled set $D_l$, while all the remaining samples constitute the unlabeled set $D_u$. We provides the dataset statistics in Table \ref{tab:datasets}.

\paragraph{Evaluation protocol.}
Following previous work~\cite{vaze2022generalized}, the clustering accuracy is used to evaluate the model performance, which can be represented as follows:

\begin{equation}
     ClusterA C C=\max _{p \in \mathcal{P}\left(\mathcal{Y}_u\right)} \frac{1}{N} \sum_{i=1}^N \mathds{1}\left\{\tilde{y}_i=p\left(\overline{y}_i\right)\right\},
\end{equation}
where $N$ is the number of samples in the unlabeled set $D_u$. $\tilde{y}_i$ and $\overline{y}_i$ represent ground truth label and clustering assignment label respectively. $\mathcal{P}\left(\mathcal{Y}_u\right)$ is all the permutations on $\mathcal{Y}_u$, which can be computed with Hungarian algorithm \cite{kuhn1955hungarian}.


\paragraph{Implementation details}
Following previous works~\cite{vaze2022generalized,wen2023parametric}, we use the ViT-B-16 backbone pre-trained by DINO~\cite{caron2021emerging}. For the feature encoder, we only fine-tune the final block. In the initial stage, we train the SimGCD with LSP for 200 epochs for the initial anchor selection. For all stages, we set the perturbation magnitude $\rho$ to 0.05. For all datasets, we set $k$ to 0.5$|\mathcal{C}_j|$. Since the value of $\eta$ is jointly determined by $\omega$ and $\gamma$, we set $\omega$ and $\gamma$ dynamically such that the value of $\eta$ is roughly equal to $N/|\mathcal{Y}_l|$ in the initial stage. For all generic datasets, we set $\alpha$ and $\beta$ to 1 and 0.8 respectively. For all fine-grained datasets, we set $\alpha$ and $\beta$ to 10 and 0.5 respectively. During the model training, following~\cite{wen2023parametric}, the batch size is set to 128 and the number of training epochs is 200. The initial learning rate is set to 0.1 and decay to 0.0001 with a cosine annealing schedule. The balance parameter $\lambda$ is set to 0.35. For the temperature values $\tau^u$ and $\tau^s$ of representation learning loss, we set them to 0.07 and 1.0 respectively. For the temperature values $\tau_s$ and $\tau_t$ of parametric classification loss, we set $\tau_s$ to 0.1. For $\tau_t$, it is set to 0.07 in the initial epoch and warmed up to 0.04 within the first 30 epochs with the cosine schedule. 

\subsection{Comparison to the State-of-the-Art}

We compare our method with numerous generalized category discovery methods, including variants of novel category discovery methods (k-means\cite{arthur2007k}, RS+\cite{Han2020automatically}, UNO+\cite{fini2021unified}), non-parametric GCD methods (GCD \cite{vaze2022generalized}, DCCL \cite{dccl}, PromptCAL \cite{zhang2023promptcal}, GPC \cite{zhao2023learning}, InfoSieve \cite{rastegar2024learn}, SPTNet \cite{wang2024sptnet}, CMS \cite{choi2024contrastive}) and parametric GCD methods (ORCA \cite{cao2022openworld}, 
SimGCD \cite{wen2023parametric},
Retraining with Selective Samples (RSS) \cite{ijcai2024p532},
$\mu$GCD \cite{vaze2024no},  LegoGCD \cite{cao2024solving}). Table \ref{tab:all_results} reports the results on all datasets. As can be seen from the table, for the main metric ``All" ACC, our method achieves the best results on four datasets (CIFAR-100, ImageNet-100, Stanford Cars, and Herbarium 19) by a significant margin and the second best results on CUB dataset. For the FGVC-Aircraft dataset, we also achieve competitive results. Overall, for the average results of the six datasets, our method achieves the best results on ``All" ACC and ``New" ACC, and the second best results on ``Old" ACC.    Specifically, compared with the baseline method SimGCD, our method achieves significant performance improvement on both ``Old" and ``New" classes on most datasets, confirming that our method achieves the overall performance gain by boosting the performance of ``Old" and ``New" classes evenly. For ``All" ACC, our method outperforms SimGCD by 3.5\% and 4.1\% on generic datasets CIFAR-100 and ImageNet-100, and achieves 8.7\% and 11.7\% performance improvements on fine-grained datasets CUB and Stanford Cars. In addition, our method also outperforms SimGCD by 5.6\% on the long-tailed dataset Herbarium 19, which well confirms the generality of our method. Compared with the parametric methods $\mu$GCD and LegoGCD, our method outperforms them on all datasets. By introducing the loss sharpness penalty and dynamic anchor selection strategy, our method directly alleviates the model from encoding trivial features, thus generating more accurate pseudo labels for learning.

\subsection{Ablation Study}

\paragraph{Impact of each component}
To investigate the effectiveness of each component of our method, we use ImageNet-100 and CUB datasets for evaluation.
As can be seen in Table \ref{tab:components}, incorporating LSP into SimGCD improves the accuracy of both ``Old" and ``New" classes. 
This is because the self-distillation framework, built on a pre-trained model, generates dynamic soft pseudo labels for both "Old" and "New" classes during self-training. Meanwhile, the LSP module penalizes the worst-case sharpness of the loss across all training data, enabling the model to effectively suppress the encoding of trivial features for both "Old" and "New" classes, thereby enhancing pseudo-label quality. In the meanwhile, DAS dynamically selects reliable anchors for the novel classes and assigns hard pseudo-labels to them, which can help the model quickly learn a more accurate feature distribution of novel classes in the early training stage, thereby significantly improving the accuracy of ``New" classes. More importantly, by combining LSP and DAS, our method further improves the ``All" ACC, confirming that LSP and DAS can simultaneously promote the model to generate more accurate pseudo labels.


\begin{table}[t]
      \small
      \centering
      \caption{Ablation study of different components.}
      \label{tab:components}
      \resizebox{0.8\columnwidth}{!}{\begin{tabular}{lcccccc}
        \toprule
        \multirow{2}[3]{*}{Method}  &
        \multicolumn{3}{c}{ImageNet-100} & \multicolumn{3}{c}{CUB}\\
        \cmidrule(lr){2-4} \cmidrule(lr){5-7}
         & All & Old & New & All & Old & New \\
        \midrule
        SimGCD & 83.0 & 93.1 & 77.9 & 60.3 & 65.6 & 57.7\\
         + LSP & 85.4 & \textbf{94.2} & 81.0  & 64.6 & \textbf{71.6} & 61.1 \\
         + DAS & 84.8 & 92.9 & 80.7 & 66.5 & 67.0 & 66.2 \\
        + LSP + DAS (Ours)  & \textbf{87.1} & 93.9 & \textbf{83.7} & \textbf{69.0} & 69.1 & \textbf{68.9} \\
        \bottomrule
      \end{tabular}}
\end{table}

\paragraph{Different training strategy} 
We performed ablations with different training strategies and the results are reported in Table \ref{tab:training_strategy}. In Table \ref{tab:training_strategy}, ``w/o pre-train" represents that the clustering results of SimGCD with LSP  are not used for initial anchor selection, and only start dynamic anchor selection at the end of warm-up training (\textit{i.e. 30 epochs}). In contrast, ``Fixed" means that only the initial anchor selection is performed and these anchors are used for subsequent training. As can be seen from the table, DAS achieves the best ``All" ACC on both datasets. This is because using the initial clustering results can provide better initial anchors for the model, which facilitates the rapid learning of the model for novel classes. As the training progresses, the model can provide better anchors, using dynamic anchor selection can avoid overfitting to the noise in the initial anchors.

\begin{table}[t]
      \small
      \centering
      \caption{Ablation study of different training strategies.}
      \label{tab:training_strategy}
      \resizebox{0.8\columnwidth}{!}{\begin{tabular}{lcccccc}
        \toprule
        \multirow{2}[3]{*}{Method}  &
        \multicolumn{3}{c}{ImageNet-100} & \multicolumn{3}{c}{CUB}\\
        \cmidrule(lr){2-4} \cmidrule(lr){5-7}
         & All & Old & New & All & Old & New \\
        \midrule
        w/o pre-train & 86.1 & \textbf{94.1} & 82.0 & 64.0 & 69.5 & 61.2 \\
        Fixed & 85.5 & 93.9 & 81.3 & 67.7 & \textbf{70.3} & 66.4 \\
        DAS & \textbf{87.1} & 93.9 & \textbf{83.7} & \textbf{69.0} & 69.1 & \textbf{68.9} \\
        \bottomrule
      \end{tabular}}
\end{table}

\paragraph{Different anchor selection strategy} We also performed ablations with different anchor selection strategies and the results are reported in Table \ref{tab:selection_strategy}. ``Max Prob." represents that selects $N/|\mathcal{Y}_l|$ anchors for each novel class based on classification probability at each training epoch. ``Max Density" represents that selects $N/|\mathcal{Y}_l|$ anchors for each novel class based on KNN feature density at each training epoch. It can be seen from Table \ref{tab:selection_strategy} that our method achieves the best ``All" ACC. 
This is because directly selecting samples based on classification probability or KNN density may capture noisy samples that the model has overfitted.
In contrast, DAS generates candidate anchor sets with high purity by selecting samples that are nearest to the density peak point, so that DAS can be less affected by noise and achieve better performance.  

\begin{table}[t]
      \small
      \centering
      \caption{Ablation study of different anchor selection strategies.}
      \label{tab:selection_strategy}
      \resizebox{0.8\columnwidth}{!}{\begin{tabular}{lcccccc}
        \toprule
        \multirow{2}[3]{*}{Method}  &
        \multicolumn{3}{c}{ImageNet-100} & \multicolumn{3}{c}{CUB}\\
        \cmidrule(lr){2-4} \cmidrule(lr){5-7}
         & All & Old & New & All & Old & New \\
        \midrule
        Max Prob. & 86.5 & \textbf{94.0} & 82.7 & 66.1 & 68.4 & 64.9 \\
        Max Density & 86.6 & \textbf{94.0} & 82.8 & 67.6 & \textbf{71.3} & 65.7 \\
        DAS & \textbf{87.1} & 93.9 & \textbf{83.7} & \textbf{69.0} & 69.1  & \textbf{68.9} \\
        \bottomrule
      \end{tabular}}
\end{table}

\subsection{Further Analyses}


\paragraph{Sensitivity analysis of hyper-parameter $\beta$} $\beta$ (first defined in Section \ref{subsec:DAS}) represents the ratio of the number of samples in the candidate anchor set to the number of samples in the corresponding cluster. We evaluate the selection of different $\beta$ values, and the results are shown in Table \ref{tab:candidate_num}. As can be seen from the table, for the CIFAR-100 dataset, the value of $\beta$ in the range of 0.6 to 0.8 can achieve stable and good results. However, for the CUB dataset, the value of $\beta$ in the range of 0.5 to 0.6 can achieve stable and good results. This is because for fine-grained datasets, the number of samples per class is small (usually less than 70). which makes it easy for the model to overfit noisy samples. In order to ensure the purity of the candidate anchor set, the $\beta$ value of the fine-grained datasets should be set lower than that of the generic datasets.

\begin{table}[t]
      \small
      \centering
      \caption{Sensitivity analysis of hyper-parameter $\beta$.}
      \label{tab:candidate_num}
      \resizebox{0.8\columnwidth}{!}{\begin{tabular}{lcccccc}
        \toprule
        \multirow{2}[3]{*}{$\beta$}  &
        \multicolumn{3}{c}{CIFAR-100} & \multicolumn{3}{c}{CUB}\\
        \cmidrule(lr){2-4} \cmidrule(lr){5-7}
         & All & Old & New & All & Old & New \\
        \midrule
        0.3 & 82.8 & 84.3  & 79.9  & 67.9 & 68.3  & 67.6 \\
        0.5  & 83.0 & 84.0  & 80.9 & \textbf{69.0} & 69.1 & \textbf{68.9} \\
        0.6 & 83.4 & 84.2 & 81.7 & 68.3 & \textbf{70.5} & 67.2 \\
        0.8  & \textbf{83.7} & \textbf{84.6} & \textbf{82.0}  & 67.6 & 69.2 & 66.8 \\
        1.0 & 83.3 & 84.1 & 81.7  & 67.3  & 69.1 & 66.4 \\
        \bottomrule
      \end{tabular}}
\end{table}

\paragraph{Sensitivity analysis of hyper-parameter $\eta$}
$\eta$ (first defined in Section \ref{subsec:DAS}) represents the number of anchors selected by DAS for each novel class in each training epoch. Since its value is determined by $\omega$ and $\gamma$, its value changes dynamically during model training. Therefore, we determine the value of $\eta$ in the initial stage by setting $\omega$ and $\gamma$. Here we evaluate the different choices of $\eta$, and the results are reported in Table \ref{tab:selected_num}. As can be seen from the table, for the CIFAR-100 dataset, stale and good results can be achieved when the value of $\eta$ is in the range of $N/|\mathcal{Y}_l|$ to 3/2 $N/|\mathcal{Y}_l|$. However, for the CUB dataset, stable and good results can be achieved when the value of $\eta$ is in the range of 1/2 $N/|\mathcal{Y}_l|$ to $N/|\mathcal{Y}_l|$. This is because the baseline accuracy of the generic dataset is significantly higher than that of the fine-grained dataset, so more anchors can be selected while ensuring high purity. Nevertheless, good results can be achieved on all types of datasets when the value of $\eta$ is set to $N/|\mathcal{Y}_l|$.

\begin{table}[t]
      \small
      \centering
      \caption{Sensitivity analysis of hyper-parameter $\eta$.}
      \label{tab:selected_num}
      \resizebox{0.8\columnwidth}{!}{\begin{tabular}{lcccccc}
        \toprule
        \multirow{2}[3]{*}{$\eta$}  &
        \multicolumn{3}{c}{CIFAR-100} & \multicolumn{3}{c}{CUB}\\
        \cmidrule(lr){2-4} \cmidrule(lr){5-7}
         & All & Old & New & All & Old & New \\
        \midrule
        1/4 $N/|\mathcal{Y}_l|$ & 82.9 & 84.6 & 79.6 & 67.8 & 69.1 & 67.2\\
        1/2 $N/|\mathcal{Y}_l|$ & 83.0 & 84.3 & 80.3 & 68.1 & 69.0 & 67.7 \\
        $N/|\mathcal{Y}_l|$ & \textbf{83.7} & \textbf{84.6} & 82.0  & \textbf{69.0} & \textbf{69.1} & \textbf{68.9} \\
        3/2 $N/|\mathcal{Y}_l|$ & 83.6 & 84.1 & \textbf{82.5} & 66.4 & 66.0 & 66.7  \\
        \bottomrule
      \end{tabular}}
\end{table}

\paragraph{Sensitivity analysis of hyper-parameter $\alpha$}
$\alpha$ (first defined in Section \ref{subsec:DAS}) is the number of epochs for training using the anchors selected in the initial stage. Here we evaluate the different choices of $\alpha$, and the results are shown in Table \ref{tab:begin_epoch}. It can be seen that for the CIFAR-100 dataset, good results can be achieved when the value of $\alpha$ is in the range of 1 to 30. However, for the CUB dataset, stable and good results can be achieved when the value of $\alpha$ is in the range of 5 to 20. This is because fine-grained tasks are usually more difficult than generic tasks, making the model need more epochs of training to obtain enough discriminative capability for effective dynamic anchor selection. At the same time, in order to prevent the model from overfitting the noise, the value of $\alpha$ should not be too large. 

\paragraph{Sensitivity analysis of hyper-parameter $k$}

\revised{For $k$ in $K$-nearest neighbor density estimate, we evaluate its different value choices, and the results are reported in Table \ref{tab:k_value}. As can be seen from the table, stable and good results can be achieved with values ranging from 0.3$|\mathcal{C}_j|$ to $|\mathcal{C}_j|$ for $k$.}

\paragraph{Sensitivity analysis of hyper-parameter $\rho$}

\revised{$\rho$ is the hyper-parameter that controls the amplitude of perturbation in LSP. Here we evaluate the different choices of $\rho$, and the results on Stanford cars and CUB datasets are shown in Table \ref{tab:perturbation_value}. As can be seen from the Table \ref{tab:perturbation_value}, when the value of $\rho$ is in the range of 0.05 to 1, stable and good results can be achieved on both datasets.}



\begin{table}[t]
      \small
      \centering
      \caption{Sensitivity analysis of hyper-parameter $\alpha$.}
      \label{tab:begin_epoch}
      \resizebox{0.8\columnwidth}{!}{\begin{tabular}{lcccccc}
        \toprule
        \multirow{2}[3]{*}{$\alpha$}  &
        \multicolumn{3}{c}{CIFAR-100} & \multicolumn{3}{c}{CUB}\\
        \cmidrule(lr){2-4} \cmidrule(lr){5-7}
         & All & Old & New & All & Old & New \\
        \midrule
        1 & 83.7 & \textbf{84.6} & 82.0 & 66.4 & 67.2 & 66.0\\
        5 & 83.6 & 84.0 & 82.8  & 68.4 & 67.8 & 68.7 \\
        10 & 83.7 & 84.1 & \textbf{83.1}  & \textbf{69.0} & \textbf{69.1} & \textbf{68.9} \\
        20 & 83.6 & 84.1 & 82.6 & 68.5 & 69.6  & 67.9 \\
        30 & \textbf{83.8} & 84.3 & 82.8 & 67.9 & 69.1  & 67.3 \\
        \bottomrule
      \end{tabular}}
\end{table}

\begin{table}[t]
      \small
      \centering
      \caption{Sensitivity analysis of hyper-parameter $k$.}
      \label{tab:k_value}
      \resizebox{0.8\columnwidth}{!}{\begin{tabular}{lcccccc}
        \toprule
        \multirow{2}[3]{*}{$k$}  &
        \multicolumn{3}{c}{CIFAR-100} & \multicolumn{3}{c}{CUB}\\
        \cmidrule(lr){2-4} \cmidrule(lr){5-7}
         & All & Old & New & All & Old & New \\
        \midrule
        0.1 $|\mathcal{C}_j|$ & 83.3 & 84.2  & 81.6  & 68.2 & 69.2 & 67.7 \\
        0.3 $|\mathcal{C}_j|$ & 83.5 & 84.5  & 81.5  & 68.5 & 70.7 & 67.5 \\
        0.5 $|\mathcal{C}_j|$  & \textbf{83.7} & \textbf{84.6} & 82.0 & \textbf{69.0} & 69.1 & \textbf{68.9} \\
        0.8 $|\mathcal{C}_j|$  & 83.5 & 84.3 & 81.8 & 68.5 & 70.9 & 67.3 \\
        $|\mathcal{C}_j|$ & 83.5 & 84.1  & \textbf{82.2}  & 68.4  & \textbf{71.3} & 66.9 \\
        \bottomrule
      \end{tabular}}
\end{table}

\begin{table}[t]
      \small
      \centering
      \caption{Sensitivity analysis of hyper-parameter $\rho$.}
      \label{tab:perturbation_value}
      \resizebox{0.8\columnwidth}{!}{\begin{tabular}{lcccccc}
        \toprule
        \multirow{2}[3]{*}{$\rho$}  &
        \multicolumn{3}{c}{Stanford Cars} & \multicolumn{3}{c}{CUB}\\
        \cmidrule(lr){2-4} \cmidrule(lr){5-7}
         & All & Old & New & All & Old & New \\
        \midrule
        0.02 & 63.3 & \textbf{80.2} & 55.1 & 67.3 & 68.1 & 67.0 \\
        0.05 & \textbf{65.5} & 79.7 & \textbf{58.8} & 69.0 & 69.1 & \textbf{68.9} \\
        0.1 & 64.6 & 77.4 & 58.5  & \textbf{69.1} & 70.4 & 68.4 \\
        0.15 & 62.8 & 77.5 & 55.8 & 68.7 & 70.4  & 67.8 \\
        0.25 & 62.4 & 77.1 & 55.3  & 67.4 & \textbf{70.9}  & 65.7 \\
        0.5 & 60.6 & 76.8 & 52.9 & 67.0  & 70.2 & 65.3 \\
        \bottomrule
      \end{tabular}}
\end{table}

\paragraph{The results with estimated class number}
\begin{table}[t]
\small
    \centering
    \caption{Results with the estimated number of classes}
     \resizebox{\columnwidth}{!}{
\begin{tabular}{lccccccc}
\toprule
\multirow{2}[3]{*}{Method}                                   & \multirow{2}[3]{*}{Known $C$} &\multicolumn{3}{c}{CUB} &\multicolumn{3}{c}{Stanford Cars} \\
 \cmidrule(lr){3-5} \cmidrule(lr){6-8}
& & All  & Old  & New  & All  & Old  & New \\
\midrule
GCD~\cite{vaze2022generalized}& GT (200/196) & 51.3& 56.6& 48.7 &39.0 &57.6& 29.9  \\
SimGCD~\cite{wen2023parametric}      & GT (200/196) & 60.3&	{65.6}&	57.7 & 53.8& 71.9 &45.0  \\
$\mu$GCD~\cite{vaze2024no} & GT (200/196) &65.7& 68.0& 64.6& 56.5& 68.1 &50.9\\
\textbf{Ours}                     & GT (200/196) &
\textbf{69.0}&\textbf{69.1} &\textbf{68.9} & \textbf{65.2} & \textbf{79.7}& \textbf{58.8}\\
\midrule
GCD~\cite{vaze2022generalized}& Est. (231/230) & 47.1& 55.1 &44.8 &35.0 &56.0& 24.8 \\
SimGCD~\cite{wen2023parametric}                 & Est. (231/230) & 61.5 & 66.4 & 59.1 & 49.1 & 65.1  &41.3\\
$\mu$GCD~\cite{vaze2024no} &  Est. (231/230) &
62.0& 60.3 &62.8& 56.3& 66.8& 51.1 \\

\textbf{Ours}            & Est. (231/230)   & \textbf{66.6} &\textbf{67.0} &\textbf{66.4} &\textbf{64.4} &\textbf{77.8} &\textbf{57.9} \\
\bottomrule
\end{tabular}}
\label{tab:est_c}
\end{table}

In the real world setting, the number of novel classes is usually not available. To evaluate the performance of our method in this more challenging scenario, following $\mu$GCD, we use an off-the-shelf method to estimate the number of categories in the CUB and Stanford cars datasets and conduct the model training. The comparison results are shown in Table \ref{tab:est_c}. As can be seen from the table, our method not only achieves the best results, but also shows a significant performance improvement over SimGCD, which well confirms the effectiveness of our method in the real world setting.   

\paragraph{The results with DINOv2 backbone} Following SPTNet, we also evaluated the performance of our method on the more powerful DINOv2 backbone, and the comparison results are shown in Table \ref{tab:dinov2}. As can be seen from the table, our method still achieves the best results, confirming the effectiveness of our method under the more powerful pre-training representation.

\paragraph{Analysis of Different Fine-tuning Strategies} \revised{We conducted experiments of fine-tuning different numbers of ViT blocks on the CIFAR-100
and CUB datasets, and the experimental results are presented in Table \ref{tab:blocks}. It can be seen
from the table that as the number of fine-tuned blocks increases (\textit{i.e.} the number of frozen blocks decreases), the performance of the
model gradually declines. This is because the features in the middle-level blocks mainly focus
on forming object parts and modeling the relationships between these parts, which is very
important for cross-class generalization. Fine-tuning these blocks will cause the model to lose
a large number of cross-class generalization feature structures. This leads to the decline of
the model’s clustering performance. However, for the case of fine-tuning different blocks, our
method can still consistently improve the model performance, confirming the adaptability of
the proposed method.}

\begin{table}[t]
      \small
      \centering
      \caption{The results with DINOv2 backbone.}
      \label{tab:dinov2}
      \resizebox{0.8\columnwidth}{!}{\begin{tabular}{lcccccc}
        \toprule
        \multirow{2}[3]{*}{Method}  &
        \multicolumn{3}{c}{ImageNet-100} & \multicolumn{3}{c}{CUB}\\
        \cmidrule(lr){2-4} \cmidrule(lr){5-7}
         & All & Old & New & All & Old & New \\
        \midrule
        SimGCD & 88.5 & 96.2 & 84.6  & 74.9 & 78.5 & 73.1 \\
        SPTNet  & 90.1 & 96.1 & 87.1 & 76.3 & 79.5 & 74.6 \\
        \midrule
        Ours & \textbf{92.0} & \textbf{96.8}  & \textbf{89.5}  & \textbf{79.8}  & \textbf{81.7} & \textbf{78.8} \\
        \bottomrule
      \end{tabular}}
\end{table}

\begin{table}[t]
      \small
      \centering
      \caption{Analysis of different fine-tuning
strategies. ``(number)'' represents the number of frozen ViT blocks.}
      \label{tab:blocks}
      \resizebox{0.8\columnwidth}{!}{\begin{tabular}{lcccccc}
        \toprule
        \multirow{2}[3]{*}{Method}  &
        \multicolumn{3}{c}{CIFAR-100} & \multicolumn{3}{c}{CUB}\\
        \cmidrule(lr){2-4} \cmidrule(lr){5-7}
         & All & Old & New & All & Old & New \\
        \midrule
        SimGCD (8) & 75.5 & 79.4 & 67.7  & 53.0 & 53.7 & 52.7 \\
        Ours (8) & \textbf{78.0} & \textbf{80.7} & \textbf{72.5}  & \textbf{58.6} & \textbf{59.6} & \textbf{58.1} \\
        SimGCD (9) & 76.6 & 80.2 & 69.3  & 57.8 & 60.3 & 56.5 \\
        Ours (9) & \textbf{79.1} & \textbf{82.4} & \textbf{72.6}  & \textbf{62.2} & \textbf{61.9} & \textbf{62.4} \\
        SimGCD (10) & 77.8 & 79.7 & 73.9 & 58.7 & 59.3  & 58.3 \\
        Ours (10) & \textbf{81.2} & \textbf{83.5} & \textbf{76.7}  & \textbf{64.3} & \textbf{62.2} & \textbf{65.4} \\
        SimGCD (11) & 80.1 & 81.2 & 77.8 & 60.3  & 65.6  & 57.7  \\
        Ours (11) & \textbf{83.7} & \textbf{84.6} & \textbf{82.0}  & \textbf{69.0}  & \textbf{69.1} & \textbf{68.9} \\
        \bottomrule
      \end{tabular}}
\end{table}

\paragraph{Anchor selection analysis} We analyzed the selected anchor points and compared with the multi-stage method RSS\cite{ijcai2024p532}. The comparison results are reported in Table \ref{tab:anchor}. It can be seen from the table that the anchor points selected by our method are slightly higher than RSS in quantity and significantly higher in purity, which confirm the superiority of our method. \revised{In addition, compared with ``SimGCD + DAS", introducing an additional loss sharpness penalty term in the dynamic anchor selection process can help the model select more accurate novel class anchor points, which further verifies the complementarity of LSP and DAS.}

\paragraph{The analysis of hard pseudo-labeling strategy} \revised{The purpose of hard pseudo-labeling is to encourage the model to quickly learn the general distribution of novel classes. Therefore, the purity of the anchor points used for hard pseudo-labeling must be guaranteed. In Table \ref{tab:hard_pseudo_analysis}, ``SimGCD + LSP + All Clean Anchor'' represents the clustering results assuming that all anchor points are clean. It can be seen that compared with ``SimGCD + LSP'' which completely uses soft pseudo-labels as supervision information, model training with a mixture of correct hard and soft pseudo-labels can lead to significant model performance gains.  ``SimGCD + LSP + Fixed Anchor" indicates that only the anchor points selected in the initial stage are used throughout the entire training process. It can be seen that compared with ``SimGCD + LSP", the performance improvement of ``SimGCD + LSP + Fixed Anchor" on the Imagenet-100 dataset is very limited. This indicates that the clustering performance of the model is indeed affected by the purity of anchor points. ``SimGCD+LSP+DAS" indicates that the anchor points are updated in each epoch. This dynamic anchor point selection strategy gradually improves the purity of anchor points as the training progresses (92\% for imagenet-100 and 79\% for CUB) and enhances the clustering performance. This confirms that our method can select anchor points of higher purity for hard pseudo-labeling, thereby effectively improving the model performance.}

\begin{table}[t]
\centering
\small
    \caption{Anchor selection analysis.``clean" refers to the number of clean samples selected. ``overall" refers to the number of all samples selected. ``prec" refers to percentage. }
    \label{tab:anchor}
    \resizebox{\columnwidth}{!}{%
    \begin{tabular}{cccccc}
    	\toprule
    	 Dataset & Method & All & Old & New & clean/overall (prec) \\
    	\midrule
     \multirow{3}{*}{CIFAR-100} 
    	& RSS \cite{ijcai2024p532} & 80.7
    & 81.4
    & 79.1
    & 4326/5000 (87\%)  \\
        & \revised{SimGCD + DAS} 
    & \revised{82.0}
    & \revised{82.2}
    & \revised{81.8}
    & \revised{4560/4920 (93\%)} \\
        
    	& SimGCD + DAS +LSP (Ours)
    & \textbf{83.7}
    & \textbf{84.6}
    & \textbf{82.0}
    & 5003/5200 (96\%) \\
        
    \midrule
    \midrule
    \multirow{3}{*}{CUB} 
    	& RSS \cite{ijcai2024p532}
    
    & 65.1
    & 66.2
    & 64.5
    & 1015/1386 (73\%)  \\
        & \revised{SimGCD + DAS} 
    
    & \revised{66.5}
    & \revised{67.0}
    & \revised{66.2}
    & \revised{1116/1468 (76\%)} \\
        
    	& SimGCD + DAS +LSP (Ours)
   
    & \textbf{69.0}
    & \textbf{69.1}
    & \textbf{68.9}
    & 1140/1443 (79\%) \\
     
     \bottomrule
    \end{tabular}
    }
\end{table}

\begin{table}[t]
\centering
\small
    \caption{Hard pseudo-labeling strategy analysis.``clean" refers to the number of clean samples selected. ``overall" refers to the number of all samples selected. ``prec" refers to percentage. }
    \label{tab:hard_pseudo_analysis}
    \resizebox{\columnwidth}{!}{%
    \begin{tabular}{cccccc}
    	\toprule
    	 Dataset & Method & All & Old & New & clean/overall (prec) \\
    	\midrule
     \multirow{4}{*}{ImageNet-100} 
    	& SimGCD + LSP 
    & 85.4
    & 94.2  
    & 81.0
    & -/- (-) \\
        & SimGCD + LSP + Fixed Anchor 
    & 85.5
    & 93.9
    & 81.3
    & 26759/29550 (91\%) \\
        
    	& SimGCD + LSP + DAS (Ours)
    & 87.1
    & 93.9
    & 83.7
    & 39123/42661 (92\%) \\
    &  SimGCD + LSP + All Clean Anchor
    & \textbf{88.4}
    & \textbf{94.0}
    & \textbf{85.6}
    & 39123/39123 (100\%) \\
        
    \midrule
    \midrule
    \multirow{4}{*}{CUB} 
    	& SimGCD + LSP 
     & 64.6
    & 71.6
    & 61.1
    & -/- (-) \\
        & SimGCD + LSP + Fixed Anchor 
    & 67.7
    & 70.3
    & 66.4
    & 1000/1313 (76\%) \\
        
    	& SimGCD + LSP + DAS (Ours)
    & 69.0
    & 69.1
    & 68.9
    & 1140/1443 (79\%) \\
    &  SimGCD + LSP + All Clean Anchor
    & \textbf{72.0}
    & \textbf{70.4}
    & \textbf{72.8}
    & 1140/1140 (100\%) \\
     
     \bottomrule
    \end{tabular}
    }
\end{table} 

\paragraph{Time efficiency analysis}

\revised{In Table \ref{tab:time}, we provide the running time on imagenet-100 and the average running time on three fine-grained datasets (including CUB, Stanford Cars, and FGVC-Aircraft) of the SSB benchmark. It can be seen that compared with SimGCD, ``SimGCD + LSP'' and ``SimGCD + DAS'' significantly improve the clustering performance, while the increased training time is acceptable.  Compared with recently proposed SPTNet, ``SimGCD + DAS'' achieves comparable results while its time consumption is significantly lower than that of SPTNet. Meanwhile, by combining LSP, ``SimGCD + LSP + DAS'' has significantly better performance than SPTNet while the time consumption is similar to that of SPTNet, confirming the superiority of our method. In addition, since our method does not contain any additional inference components compared with SimGCD, the best ACC can be achieved while maintaining reasonable reasoning efficiency.}


\begin{table}[h]
\small
    \centering
    \caption{Time efficiency analysis. We show ``All" accuracy, total training time, and inference time.}
\setlength{\tabcolsep}{2.5pt}      
     \resizebox{\columnwidth}{!}{
\begin{tabular}{lcccccc}
\toprule
\multirow{2}[3]{*}{Method}     
&\multicolumn{3}{c}{ImageNet-100} &\multicolumn{3}{c}{SSB}
\\
 \cmidrule(lr){2-4} \cmidrule(lr){5-7} 
& ACC  & Training  & Inference & ACC  & Training  & Inference \\
\midrule
SimGCD~\cite{wen2023parametric}  & 83.0  & 44.6h  & 591s  & 56.1 & 3.6h  & 17s \\
SPTNet~\cite{wang2024sptnet}   &  85.4 & 134.2h &601s& 61.4 &8.9h &17s  \\
SelEX~\cite{RastegarECCV2024}  & 83.1& 95.5h & 859s & 63.1& 5h &61s \\
SimGCD + LSP  & 85.4  &84.1h  & 591s  & 60.4 & 6.7h  & 17s \\
SimGCD + DAS  & 84.8  &76.8h  & 591s  & 61.7 & 6.1h  & 17s \\
SimGCD + LSP + DAS (Ours)  & \textbf{87.1}  & 131h  & 591s  & \textbf{63.4} & 9.2h  & 17s \\
\bottomrule
\end{tabular}}
\label{tab:time}
\vspace{-1em}
\end{table}


\begin{figure}[h]
    \centering
    \includegraphics[width=0.99\columnwidth]{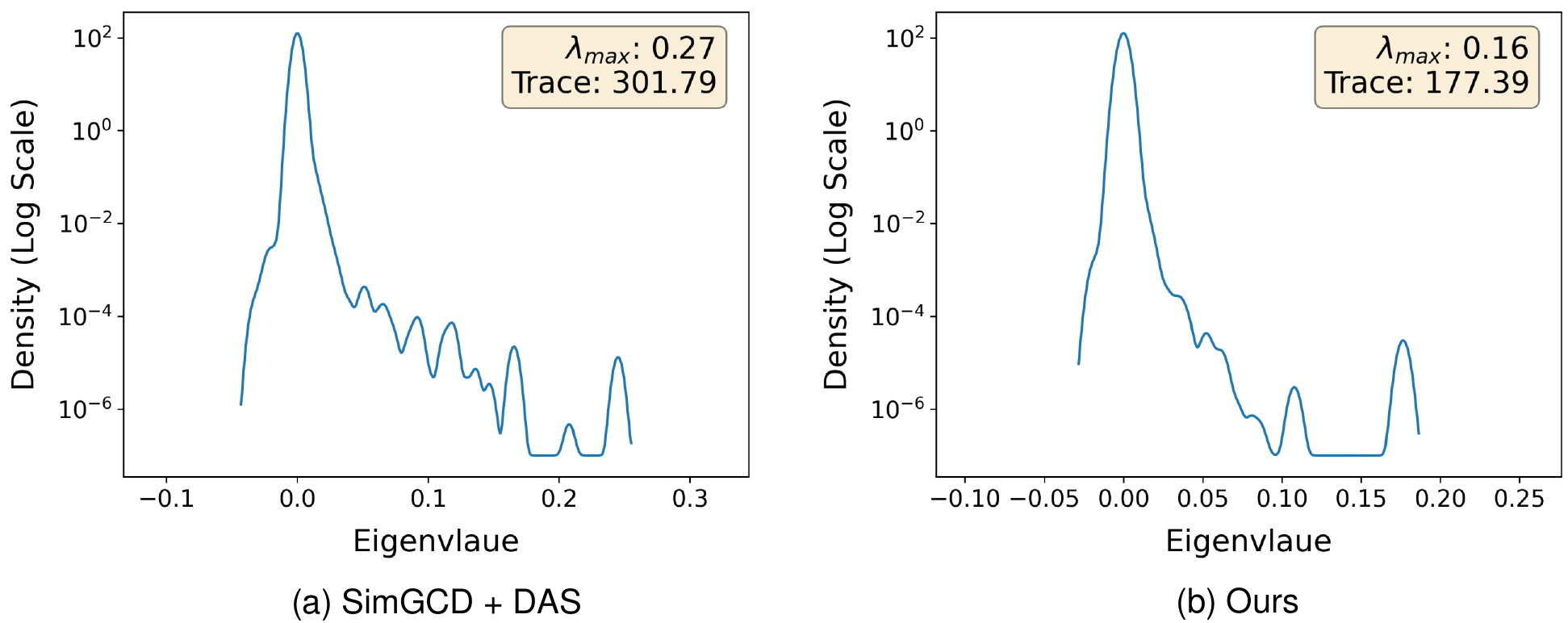}
    \caption{Hession eigenvalue distribution for ``SimGCD+DAS" and Ours, ``$\lambda_{max}$" and ``Trace" represent the maximum eigenvalue and the trace of hessian matrix, respectively. Lower ``$\lambda_{max}$" and ``Trace" means a flatter loss landscape.}
    \label{fig:hession}
\end{figure}

\begin{figure}[h]
    \centering
    \includegraphics[width=0.99\columnwidth]{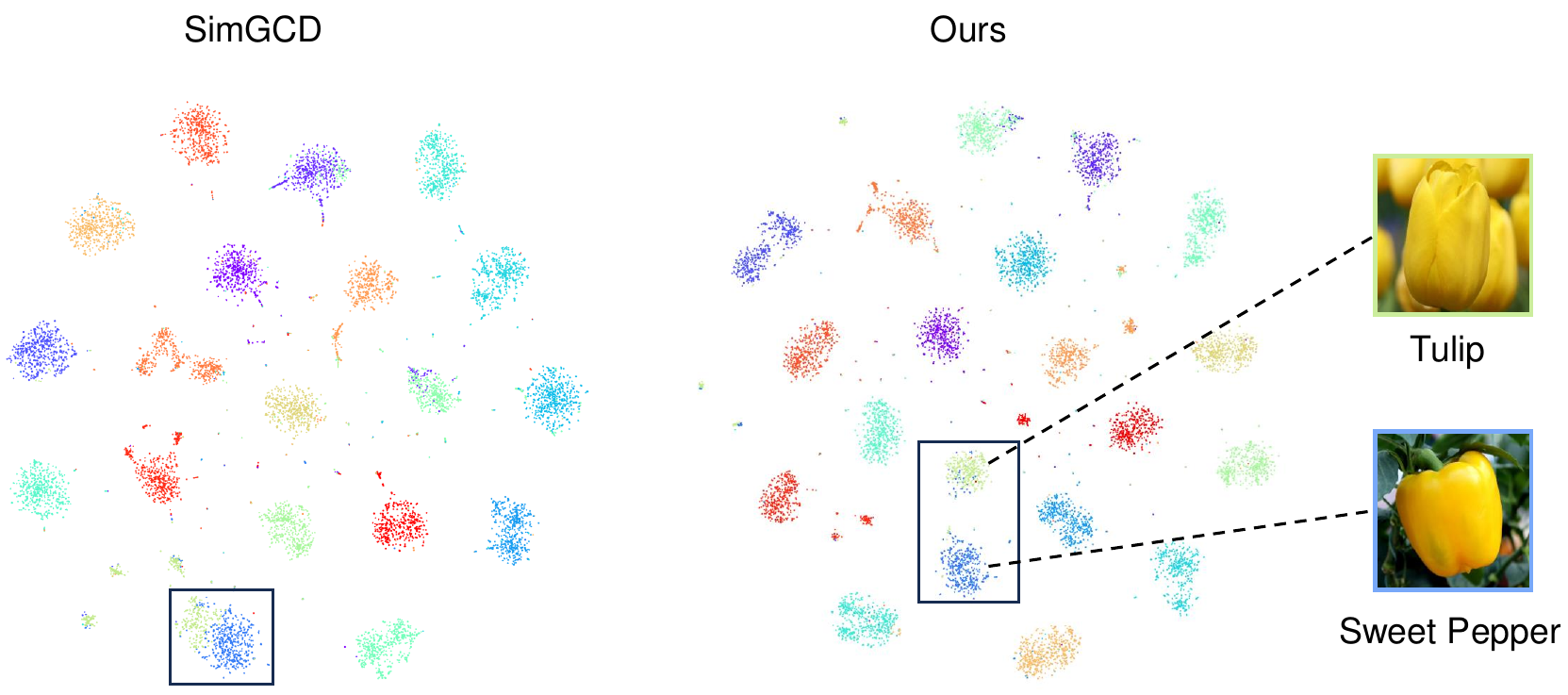}
    \vspace{-1em}
    \caption{t-SNE visualization of the representations of all novel classes of CIFAR-100.}
    \vspace{-1em}
    \label{fig:t-sne}
\end{figure}

\paragraph{Hessian Eigen Spectral Density}
 We plot hession  eigenvalue distributions in Fig.\ref{fig:hession}, as shown in Fig.\ref{fig:hession}, compared with ``SimGCD+DAS", our method has significantly smaller maximum eigenvalue (0.27 v.s. 0.16) and the trace of the Hessian matrix (301.79 v.s. 177.39), empirically demonstrating that our method can lead to a flatter minima.

\paragraph{t-SNE visualization}
 As shown in Fig. \ref{fig:t-sne}, we use t-SNE \cite{van2008visualizing} to visualize the embedding representations of SimGCD and our method on all the 20 novel classes samples of CIFAR-100 dataset. As can be seen in Fig. \ref{fig:t-sne}, compared with SimGCD, our method can effectively distinguish easily entangled semantic categories and establish clear category decision boundaries in the embedding space.

\section{Conclusions}
In this paper, we propose a novel method to alleviate the overfitting of parameterized GCD methods to noisy samples. Specifically, we look for the worst-case failure case by imposing small perturbations to the model, and then apply a worst-case loss sharpness penalty (LSP) to the model. By penalizing loss sharpness, the model can generate a flatter loss surface and effectively suppress overfitting to noisy samples. In addition, we also propose a novel dynamic anchor selection (DAS) strategy, which dynamically selects representative samples for unknown classes and assigns hard pseudo-labels to them. By combining LSP with DAS, the performance of the model can be further improved. Extensive experiments on both general datasets and fine-grained datasets confirm that our method can achieve significant improvements over the baseline method and achieve state-of-the-art results on multiple datasets. 
\bibliographystyle{IEEEtran}

\bibliography{reference}
\vfill

\end{document}